\documentclass[lettersize,journal]{IEEEtran}
\usepackage{amsmath,amsfonts}
\usepackage{algorithmic}
\usepackage{array}
\usepackage[caption=false,font=normalsize,labelfont=sf,textfont=sf]{subfig}
\usepackage{textcomp}
\usepackage{stfloats}
\usepackage{url}
\usepackage{verbatim}
\usepackage{cite}
\usepackage[pagebackref,breaklinks,colorlinks]{hyperref}
\usepackage{graphicx}
\usepackage{bbding}
\usepackage{booktabs}
\usepackage{multirow}
\usepackage{float}
\def\BibTeX{{\rm B\kern-.05em{\sc i\kern-.025em b}\kern-.08em
    T\kern-.1667em\lower.7ex\hbox{E}\kern-.125emX}}
\usepackage{balance}

\usepackage{color,xcolor}
\usepackage{caption}
\usepackage{algorithm}
\usepackage[capitalize]{cleveref}

\usepackage[export]{adjustbox}
\crefname{section}{Sec.}{Secs.}
\Crefname{section}{Section}{Sections}
\Crefname{table}{Table}{Tables}
\crefname{table}{TABLE}{TABLES}

\newcommand\blfootnote[1]{%
\begingroup
\renewcommand\thefootnote{}\footnote{#1}%
\addtocounter{footnote}{-1}%
\endgroup
}

\begin{document}
\title{DIffSteISR: Harnessing Diffusion Prior for Superior Real-world Stereo Image Super-Resolution}
\author{Yuanbo~Zhou,
        Xinlin~Zhang,
        Wei~Deng, 
        Tao~Wang,
        Tao~Tan, 
	Qinquan~Gao 
	and Tong~Tong \Envelope
}

\markboth{Journal of \LaTeX\ Class Files,~Vol.~18, No.~9, September~2020}%
{How to Use the IEEEtran \LaTeX \ Templates}

\maketitle

    \blfootnote{Yuanbo Zhou, is with the College of Physics and Information Engineering, Fuzhou University, Fuzhou, 350108, China. (e-mail: \url{webbozhou@gmail.com})}%
    \blfootnote{Xinlin Zhang, is with the College of Physics and Information Engineering, Fuzhou University, Fuzhou, 350108, China. (e-mail: \url{xinlin@fzu.edu.cn})}
    \blfootnote{Wei Deng, is with the Imperial Vision Technology, Fuzhou, 350108, China. (e-mail: \url{weideng.chn@gmail.com})}
    \blfootnote{Tao Wang, is with the College of Physics and Information Engineering, Fuzhou University, Fuzhou, 350108, China. (e-mail: \url{ortonwangtao@gmail.com})}
    \blfootnote{Tao Tan, Faulty of Applied Science, Macao Polytechnic University, Macao, 999078, China. (e-mail: \url{taotanjs@gmail.com})}
    \blfootnote{Qinquan Gao, is with the College of Physics and Information Engineering, Fuzhou University, Fuzhou, 350108, China. (e-mail: \url{gqinquan@fzu.edu.cn})}
    \blfootnote{Tong Tong, is with the College of Physics and Information Engineering, Fuzhou University, Fuzhou, 350108, China. (email: \url{ttraveltong@gmail.com})}
\begin{abstract}
We introduce DiffSteISR, a pioneering framework for reconstructing real-world stereo images. DiffSteISR utilizes the powerful prior knowledge embedded in pre-trained text-to-image model to efficiently recover the lost texture details in low-resolution stereo images. Specifically, DiffSteISR implements a time-aware stereo cross attention with temperature adapter (TASCATA) to guide the diffusion process, ensuring that the generated left and right views exhibit high texture consistency thereby reducing disparity error between the super-resolved images and the ground truth (GT) images. Additionally, a stereo omni attention control network (SOA ControlNet) is proposed to enhance the consistency of super-resolved images with GT images in the pixel, perceptual, and distribution space. Finally, DiffSteISR incorporates a stereo semantic extractor (SSE) to capture unique viewpoint soft semantic information and shared hard tag semantic information, thereby effectively improving the semantic accuracy and consistency of the generated left and right images. Extensive experimental results demonstrate that DiffSteISR accurately reconstructs natural and precise textures from low-resolution stereo images while maintaining a high consistency of semantic and texture between the left and right views.
\end{abstract}

\begin{IEEEkeywords}
Stereo Image Super-Resolution,  Diffusion Model, Texture Consistency, Reconstructing, ControlNet.
\end{IEEEkeywords}

\section{Introduction}
\IEEEPARstart{R}{eal-world} stereo image super-resolution (Real-SteISR) is a challenging task aimed at reconstructing high-quality stereo images from low-quality stereo images in the wild. Different from previous classic stereo image super-resolution works~\cite{chu2022nafssr,song2020stereoscopic,wang2019learning,wang2021symmetric,ying2020stereo} that focused only on single degradation types (e.g., Bicubic), Real-SteISR needs to handle complex degradations such as noise, blur, and other unknown real-world image characteristics. Moreover, unlike real-world single-image super-resolution (Real-ISR), Real-SteISR must consider not only the quality of the reconstructed images but also the consistency of textures and semantics between the left and right views. Additionally, the disparity of the reconstructed images should not significantly differ from the GT images.

Over the past few years, Real-ISR methods based on generative adversarial networks (GANs)~\cite{goodfellow2020generative} have improved visual perceptual quality~\cite{mou2022metric,zhang2021designing,ji2020real,ignatov2017dslr,wang2021real}. Compared to methods that only use pixel loss, GAN-based methods alleviate excessive smoothing. However, due to the instability of GAN training, manually meticulously adjusting the discriminator's structure and the weights of adversarial loss is required to avoid mode collapse and visual artifacts. Although Liang et al.~\cite{liang2022details} attempt to suppress visual artifacts by introducing additional locally discriminative learning loss, the inherent limitations of GANs still fail to generate satisfactory texture details.

\begin{figure}
\centering   
\includegraphics [width=0.35\textwidth]{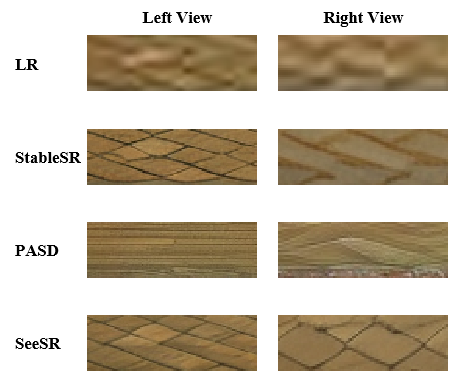}  
\caption{The visual results of the state-of-the-art Real-ISR methods based on diffusion mode for processing stereo images.}
\label{fig:1} 
\end{figure}

Recently, the success of denoising diffusion probabilistic models~\cite{ho2020denoising} in image and video generation~\cite{liu2024sora,ramesh2022hierarchical,rombach2022high,podell2023sdxl} has demonstrated the potential of diffusion model (DM) in content generation. Consequently, researchers have begun to introduce DM into the field of Real-ISR, achieving notable progress with works such as StableSR\cite{wang2024exploiting}, DiffBIR~\cite{lin2023diffbir}, PASD~\cite{yang2023pixel}, SeeSR~\cite{wu2024seesr}, SUPIR~\cite{yu2024scaling}, and 
PromptSR~\cite{chen2023image}. These methods utilize the diffusion priors of pre-trained text-to-image models to enhance image texture details. However, these methods are limited to single-image processing and are ineffective for stereo images. When applied to stereo images, the inconsistency of texture and semantic between the left and right views arise, as shown in \cref{fig:1}.

To address these challenges, we propose DiffSteISR, a DM-based solution for Real-SteISR. DiffSteISR effectively reconstructs texture details in low-quality stereo images, and simultaneously maintain semantic and texture consistency between the left and right views. Specifically, DiffSteISR constrains the diffusion process by introducing a TASCATA within Dual-UNet. Additionally, to enhance the consistency between the super-resolved images and GT images in the pixel, perceptual, and distribution space, a SOA ControlNet is integrated into the DM to control the generation of stereo images. Finally, we introduce a SSE to extract unique viewpoint soft semantic information and shared hard tag semantic information, thereby effectively improving the semantic accuracy and consistency of the generated left and right images.

The contributions of this study are summarized as follows:
\begin{enumerate} 
\item To the best of our knowledge, this is the first work that introduces diffusion priors into Real-SteISR filed.
\item We propose TASCATA, which effectively guides the diffusion generation process of stereo images, ensuring high texture consistency between the reconstructed left and right views while significantly reducing disparity error between the reconstructed images and the GT images. 
\item We introduce SOA ControlNet, which enhances the consistency of the reconstructed images with GT images across pixel space, perceptual space, and distribution space. 
\item Comprehensive experiments demonstrate that DiffSteISR achieves competitive results on both synthetic and real-world datasets. 
\end{enumerate}
\section{Related Work}

\begin{figure*}[h]
    \centering
    \includegraphics [width=0.90\textwidth]{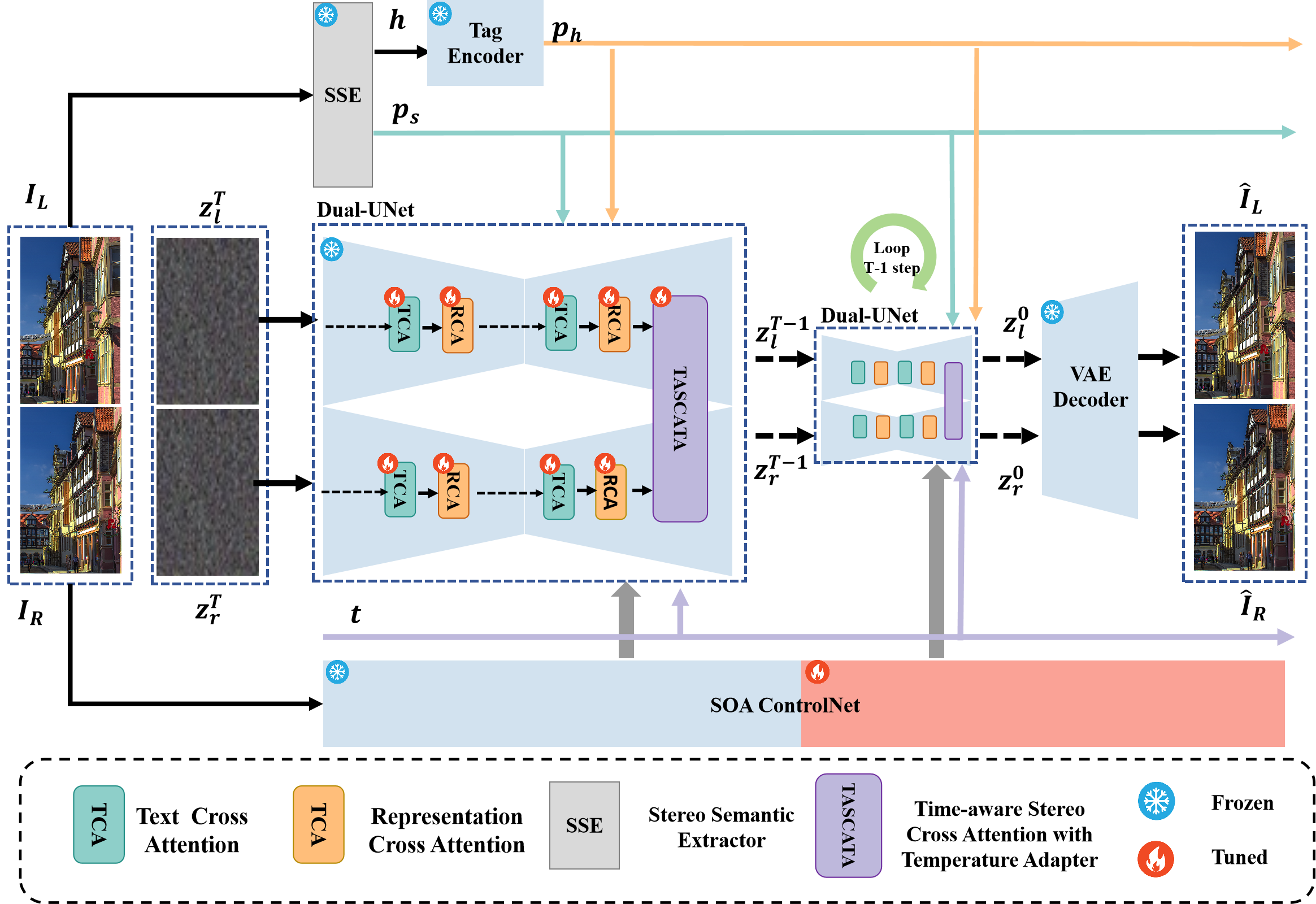}
    \caption{The framework of the proposed method consists of five parts: the stereo semantic extractor, the Tag Encoder, the SOA ControlNet, the Dual-UNet, and the VAE Decoder.}
    \label{fig:2}
\end{figure*}
\subsection{Single Image Super-Resolution Based on GANs}

Since the introduction of SRCNN~\cite{dong2015image}, the field of single image super-resolution has been revolutionized by deep learning. Over the past decade, numerous impressive methods have emerged, including VDSR~\cite{kim2016accurate}, FSRCNN~\cite{dong2016accelerating}, SRDenseNet~\cite{tong2017image}, RCAN~\cite{zhang2018image}, SwinIR~\cite{liang2021swinir}, ESRT~\cite{lu2022transformer}, and HAT~\cite{chen2023activating}, SRGAN~\cite{ledig2017photo}, ESRGAN~\cite{wang2018esrgan}, and RankSRGAN~\cite{zhang2019ranksrgan}. Researchers have gradually shifted their focus towards Real-ISR to improve image processing capabilities in practical applications. For example, Yuan et al. proposed CinCGAN~\cite{yuan2018unsupervised}, which employs a two-stage strategy to handle complex real-world images. Unlike CinCGAN, Lugmayr et al. introduced DSR~\cite{lugmayr2019unsupervised}, which first converts clean paired data to real-world settings before training the model. Similar works include FSSR~\cite{fritsche2019frequency}, GFSSR~\cite{zhou2020guided}, and RealSR~\cite{ji2020real}, all of which have achieved impressive results on the DPED ~\cite{ignatov2017dslr} dataset. However, methods like CinCGAN and DSR, along with their variants, require training a domain transform network. To address this, Zhang et al. proposed BSRGAN~\cite{zhang2021designing}, designing a more practical degradation model to replace the domain transform network, effectively improving the ability of super-resolution models to handle real-world low-quality images. Subsequently, Wang et al. introduced a second-order degradation model, RealESRGAN~\cite{wang2021real}, further enhancing the model's generalization performance. Additionally, Liang et al. proposed DASR~\cite{liang2022efficient}, which achieves high-quality real-world super-resolution by embedding degradation parameters.

\subsection{Single Image Super-Resolution Based on DMs}

To overcome the challenges of unstable training and unsatisfactory artifacts in generative adversarial networks (GANs), researchers have utilized the advantages of DMs in content generation, introducing them into the super-resolution field. For example, SR3~\cite{saharia2022image} has opened a new avenue for image super-resolution. However, this method requires retraining the entire diffusion model. To avoid this drawback, methods utilizing diffusion priors for image super-resolution have gained popularity within the research community, such as StableSR~\cite{wang2024exploiting}, DiffBIR~\cite{lin2023diffbir}, PASD~\cite{yang2023pixel}, SUPIR~\cite{yu2024scaling}, and SeeSR~\cite{wu2024seesr}. These methods primarily focus on fine-tuning a control network to assist in controlling the diffusion process with low-resolution image information, enabling the DM to generate high-quality images that closely match the content of the low-resolution images.

\subsection{Stereo Image Super-Resolution}

The field of stereo image super-resolution began to rapidly evolve following the introduction of StereoSR by Jeon~\cite{jeon2018enhancing}. Wang et al. contributed a large-scale dataset, Flickr1024~\cite{wang2019flickr1024}, providing a solid foundation for subsequent research. Researchers then shifted towards developing stereo attention modules, including the parallax attention mechanism~\cite{wang2019learning} introduced by Wang et al., the bilateral parallax attention module ~\cite{wang2021symmetric} by Wang et al., and the stereo cross attention module~\cite{chu2022nafssr} introduced by Chu et al. Furthermore, the stereo cross global learning attention module~\cite{zhou2023stereo} proposed by Zhou et al. has also improved the performance of stereo image super-resolution models. In addition to CNN-based stereo super-resolution, recent works have also explored Transformer-based approaches, such as SwinIPASSR~\cite{jin2022swinipassr} by Kai et al. Cheng et al. proposed a two-stage Transformer and CNN hybrid network~\cite{cheng2023hybrid}, achieving state-of-the-art performance.

Although the aforementioned stereo image super-resolution methods have demonstrated impressive results under known degradation conditions, similar to single image super-resolution, they often fail when dealing with real-world stereo images with unknown degradation. While some researchers have started to explore real-world stereo super-resolution, such as RealSCGLAGAN~\cite{zhou2024toward}, which achieved a milestone by combining a hybrid degradation model with an implicit discriminator, these GAN-based methods inevitably introduce artifacts and struggle to reconstruct realistic and natural textures. Building on previous work, this paper explores diffusion priors to enhance the model's ability to handle real-world stereo images with unknown degradations.

\section{Proposed Method}
\subsection{Preliminary}
In this section, we begin by introduction the principles of DMs and then provide a detailed explanation of our proposed method. DMs learn the probability distribution $p(x)$ of the data by gradually denoising a Gaussian distributed variable. They consist of a forward process and a reverse process. The forward process can be expressed as \cref{equ:dm}:
\begin{equation}
{{\cal L}_{dm}} = \mathbb{E}_{{x,t,\epsilon}}{\rm{ }}\left[ {\left. {\left\| {\epsilon  - {\epsilon _\theta }({x_t} = \sqrt {{{\overline \alpha  }_t}} x+(\sqrt {1 - {{\overline \alpha  }_t}} )\epsilon ,t)} \right\|_2^2} \right]} \right.,
\label{equ:dm}
\end{equation}

\noindent where $x$ represents data sampled from the distribution $p(x)$, \( \overline \alpha_t \) are constants, $\epsilon$ is noise sampled from a standard normal distribution, $t$ denotes the time step, $x_t$ is the noisy version of the input $x$ at time step $t$, and $\epsilon _\theta$ is the network predicting the noise $\epsilon$. Once $\epsilon _\theta$ is trained, new data that follows the distribution $p(x)$ can be generated iteratively through the reverse process. For the stable diffusion model, the training objective becomes \cref{equ:ldm}.

\begin{equation}
{{\cal L}_{ldm}} = \mathbb{E}_{{z,c,t,\epsilon }}{\rm{ }}\left[ {\left. {\left\| {\epsilon  - {\epsilon _\theta }({z_t} = \sqrt {{{\overline \alpha  }_t}} z+(\sqrt {1 - {{\overline \alpha  }_t}} )\epsilon ,c,t)} \right\|_2^2} \right]} \right.,
\label{equ:ldm}
\end{equation}

\noindent where, $z$ = $E(x)$ denotes the latent vector encoded by a variational autoencoder (VAE), and $c$ represents conditions (i.e. text prompts).
\subsection{Overview}

As shown in \cref{fig:2}, our method primarily introduces stable diffusion priors pre-trained on the Laion-5B \cite{schuhmann2022laion} text-to-image dataset to reconstruct realistic and natural textures that are missing from low-resolution images. DiffSteISR consists of five main components: the \textbf{Stereo Semantic Extractor }, the \textbf{Tag Encoder}, the \textbf{SOA ControlNet}, the \textbf{Dual-UNet}, and the \textbf{VAE Decoder}.

The SSE is responsible for extracting semantic information from the input low-resolution images. The tag encoder encodes the extracted high-level semantic information (hard tags) into semantic vectors. In particular, the tag encoder in this work uses OpenClip-ViT/H~\cite{cherti2023reproducible}. The SOA ControlNet conditionally incorporates the structure and texture information of the low-quality input image into the Dual-UNet. This ensures that the Dual-UNet enhances fidelity without altering the low-level features of the original low-resolution image.

The Dual-UNet, the most critical component, is designed to iteratively denoise the latent vectors after noise addition. It comprises two pretrained UNets with shared parameters, connected by a time-aware stereo cross attention with temperature adapter to ensure high consistency in texture between the generated left and right images. Finally, the VAE Decoder decodes the iteratively refined latent vectors from the Dual-UNet into image space. In this work, we use the default VAE Decoder from Stable Diffusion 2.0.\footnote{https://huggingface.co/stabilityai/stable-diffusion-2}

\subsection{Stereo Semantic Extractor}

\begin{figure}
\centering
\includegraphics [width=0.48\textwidth]{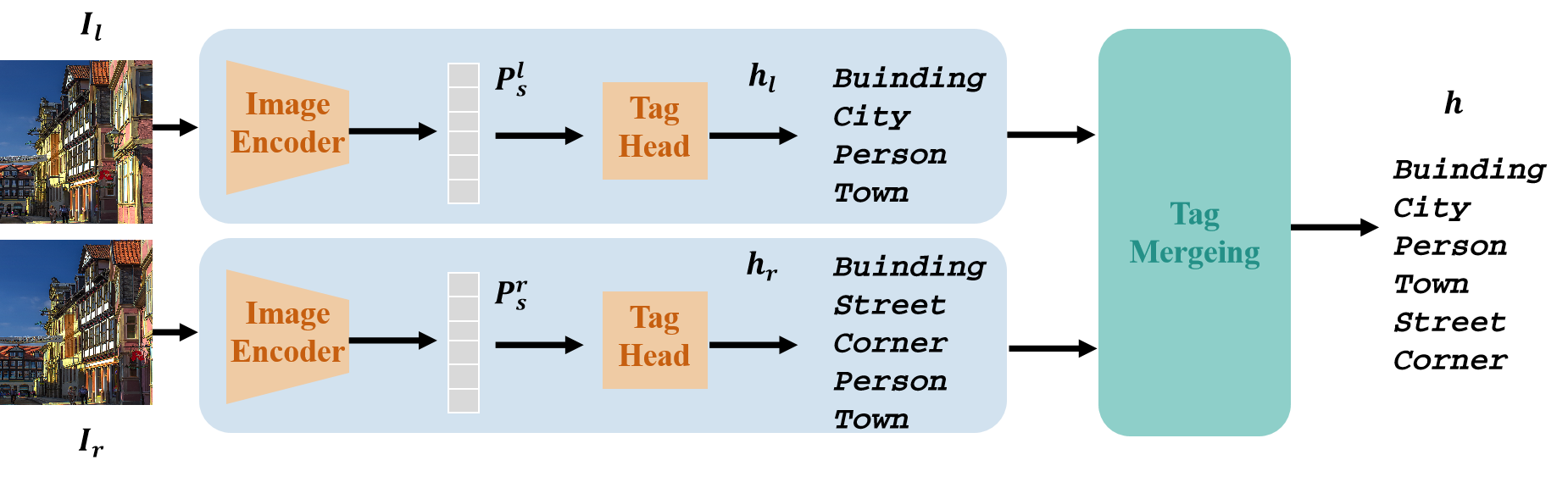}
\caption{The architecture of the stereo semantic extractor consists of an image encoder, tag head, and tag merging module.}
\label{fig:3}
\end{figure}

A series of studies such as SeeSR~\cite{wu2024seesr}, PromptSR~\cite{chen2023image}, and PASD~\cite{yang2023pixel} have demonstrated that high-quality prompts can effectively enhance the quality of generated images and reduce semantic distortions. Following this guideline, DiffSteISR employs a tag-style prompt that is suitable for image super-resolution. As shown in \cref{fig:3}, the stereo semantic extractor mainly comprises two pre-trained DAPE models~\cite{wu2024seesr} with shared parameters and a tag merging module. The extracted prompts are divided into two categories: stereo hard prompts $h$ and left/right soft prompts $p_{s}^{l}$ and $p_{s}^{r}$.

The merged tags $h$ are sent to the tag encoder to extract text embeddings $p_{h}$, effectively preventing high-level semantic differences between the generated left and right images. Additionally, the soft embeddings $p_{s}=\{p_{s}^{l},p_{s}^{r}\}$ for the left and right views are respectively injected into the Dual-UNet in~\cref{fig:2} through cross-attention, enabling the Dual-UNet to generate distinct images based on the different soft embeddings. This method overcomes the limitations of using only hard tags. The entire tags extraction process can be represented by \cref{equ:1}.

\begin{equation}
\label{equ:1} 
 \begin{aligned}
  &p_{s}^{l},p_{s}^{l}=\text{IM}({{I}_{l}}),\text{IM}({{I}_{r}}), \\ 
  &{{h}_{l}},{{h}_{r}}=\text{TH}(p_{\text{s}}^{l}),\text{TH}(p_{s}^{l}), \\ 
  &h\ =\text{TM}({{h}_{l}},{{h}_{r}}),  
 \end{aligned}
 \end{equation}

\noindent where ${I}_{r}$ and ${I}_{r}$ represent the input left and right images, IM denotes the image encoder, $p_{s}^{l}$ and $p_{s}^{r}$ represent the soft embeddings of the left and right views, TH represents the tag head, ${h}_{l}$ and ${h}_{r}$ represent the hard tags of the left and right views, and TM denotes the tag merging module. The final merged hard tag is denoted as $h$. The TM performs a set operation to remove duplicate tags and maintain consistency in hard tags between the left and right views.

\subsection{Stereo Omni Attention Control Network}
\label{sec:soacn}
In diffusion prior-based image super-resolution, it is crucial to simultaneously maintain the consistency between the generated images and ground truth (GT) images at the pixel level, perceptual level, and distributional level. Previous methods, such as StableSR~\cite{wang2024exploiting} and SeeSR~\cite{wu2024seesr}, directly fed the original low-resolution images into the control network after passing through a simple image encoder. Although straightforward, this method neglects the complementary information between stereo images, resulting in poor quality of the generated stereo images. Furthermore, for stereo image super-resolution, it is essential to reduce disparity error between the generated images and the GT images.

  \begin{figure}
    \centering
    \includegraphics [width=0.48\textwidth]{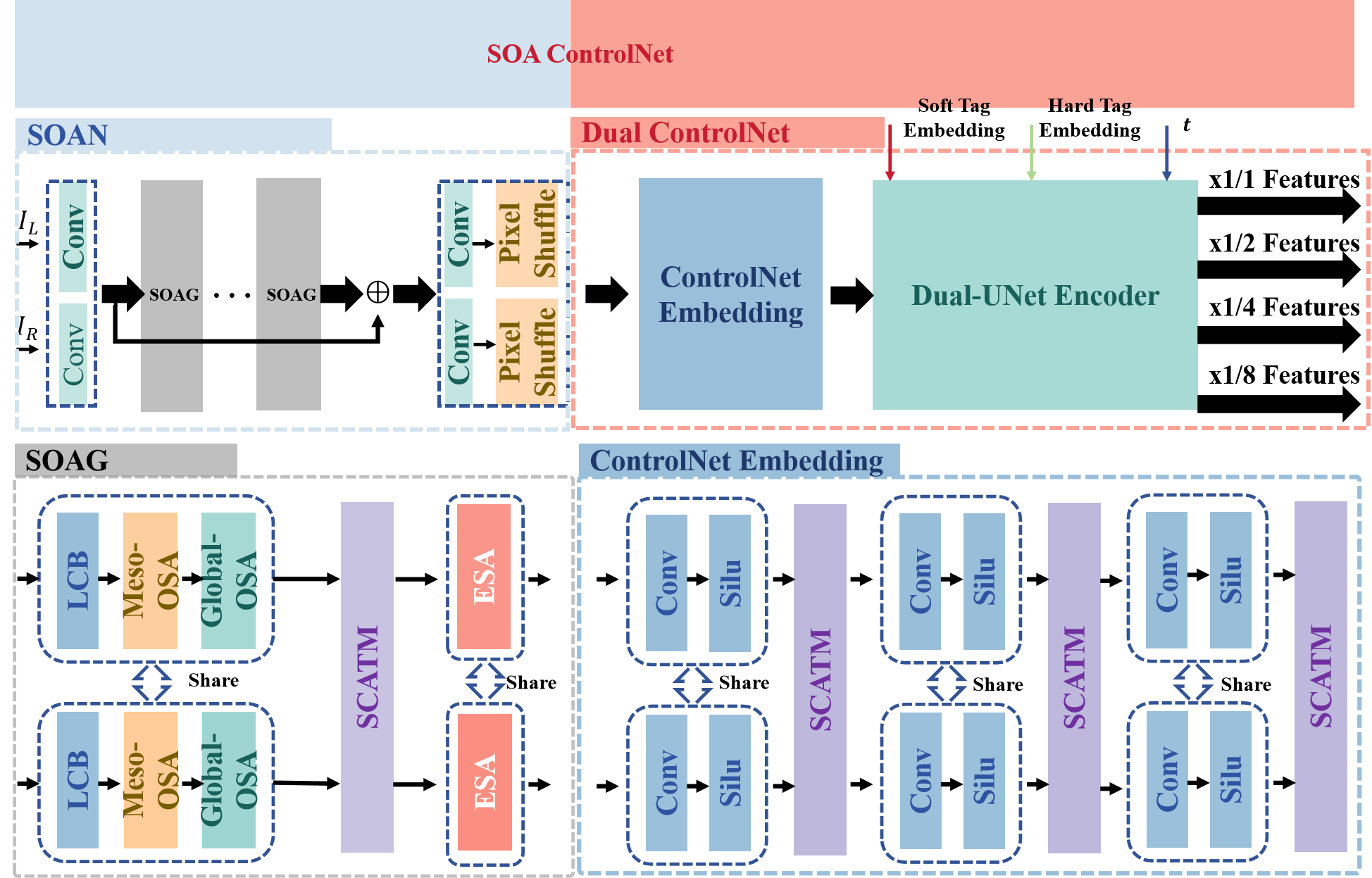}
    \caption{The architecture diagram of stereo omni attention control network.}
    \label{fig:4}
    \end{figure}

To address these problem, we propose SOA ControlNet. As shown in \cref{fig:4}, SOA ControlNet consists of two parts: the stereo omin attention network (SOAN) and the dual control network (Dual ControlNet). The SOASRN is an ultra-lightweight stereo super-resolution network composed of convolution layers, stereo omin attention group (SOAG), and pixel shuffle layer~\cite{shi2016real}. Its main function is to preprocess the low-resolution stereo images, removing partial degradation and fusing the information between the two views, providing more accurate and rich details for the stereo diffusion process, thereby facilitating the generation of high-quality details.

The Dual ControlNet consists of the Controlnet Embedding layer and the Dual-UNet Encoder. Its main purpose is to obtain control features at different scales to guide the diffusion in stereo super-resolution. Notably, the SOAN needs to be pretrained, and during the training of the diffusion stereo super-resolution, only the Dual ControlNet is trained while the SOAN is frozen.

The detailed composition of SOAG can also be found in the lower left corner of \cref{fig:4}. It consists of the local convolution block (LCB) for extracting intra-image information, Meso-OSA~\cite{wang2023omni}, Global-OSA~\cite{wang2023omni}, and ESA~\cite{liu2020residual}, as well as the SCATM~\cite{zhou2023stereo} for fusing inter-image information. The Controlnet Embedding layer comprises three shared convolution layers at different scales and SCATM~\cite{zhou2023stereo}. The detailed structure can also be obtained from the lower right corner of ~\cref{fig:4}.

\subsection{Time-aware Stereo Cross Attention with Temperature Adapter }
\begin{figure}
    \centering
    \includegraphics [width=0.48\textwidth]{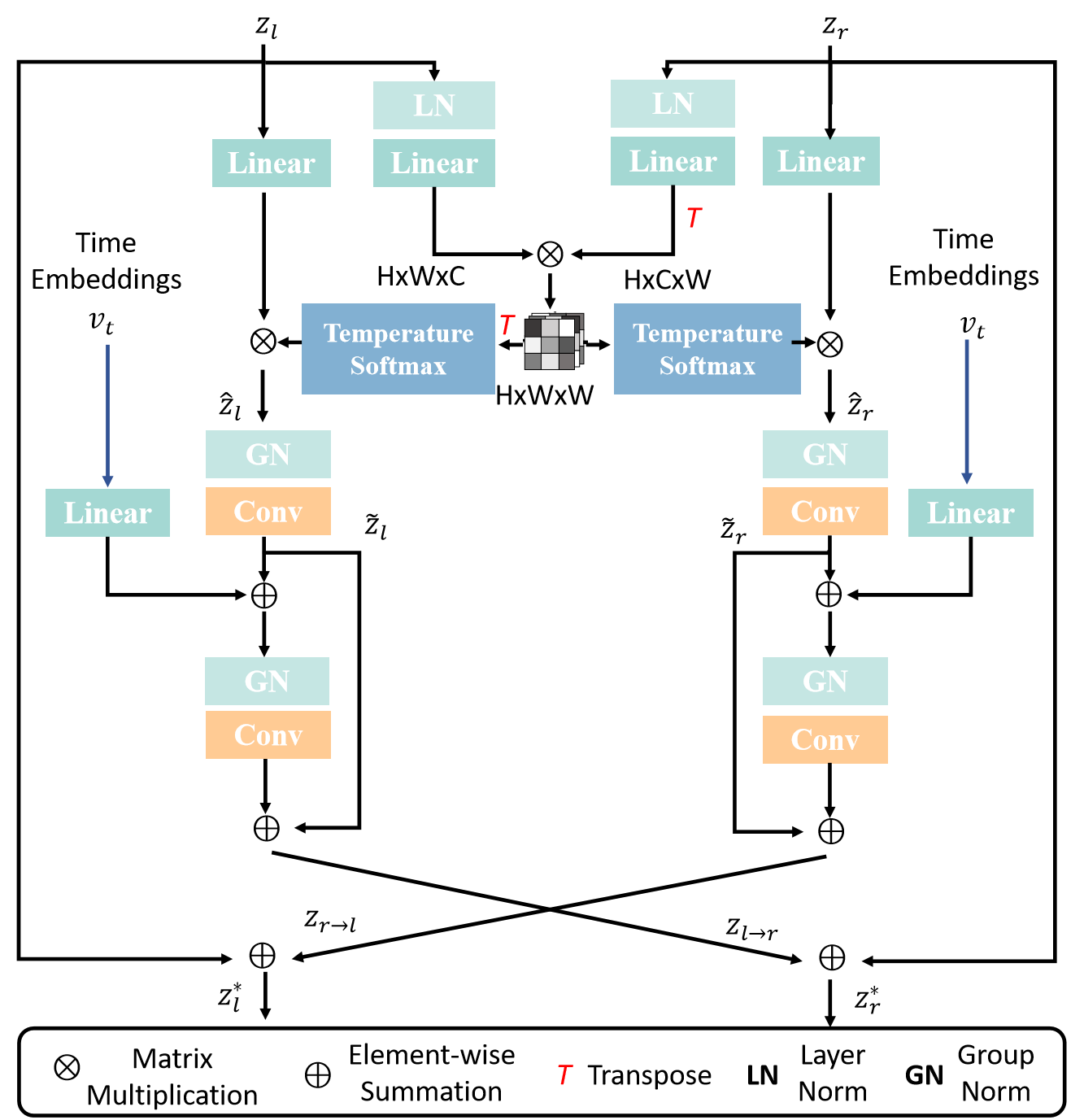}
    \caption{The architecture of time-aware stereo cross attention with temperature adapter.}
    \label{fig:5}
\end{figure}
The high consistency of texture between the left and right views is essential for accurately reconstructing depth information in 3D vision. However, due to the randomness of diffusion models, the consistency of the generated left and right views in structure and texture is often low, as shown in \cref{fig:1}. To tackle this problem, we introduce a TASCATA, which enables the diffusion model to consider both views' information during the diffusion process. As illustrated in \cref{fig:5}, the adapter consists of a dual-view cross-attention branch and a temporal embedding branch. The dual-view cross-attention branch primarily merges information from both views, constraining the Dual-Unet to consider the texture of both views, thereby enhancing the consistency of the generated left and right views. The temporal embedding branch integrates time embedding into the dual-view information fusion process, allowing temporal information to bind with features. The rationale for embedding time information into the latent space is that the noise levels of latent variables differ at different diffusion steps, embedding temporal information facilitates the adaptive fusion of dual-view information across different time steps. Specifically, given the features ${z}_{l}$ and ${z}_{r}$ of the input left and right views, and the time embedding vector ${v}_{t}$, the fused left and right feature maps can be obtained through the fusion process expressed in \cref{equ:2}:

\begin{equation}
    \label{equ:2}
    \begin{aligned}
  & {{z}_{r\to l}}={{\operatorname{TASCA}}_{r\to l}}\left( {{z}_{l}},{{z}_{r}},{{v}_{t}} \right), \\ 
 & {{z}_{l\to r}}={{\operatorname{TASCA}}_{l\to r}}\left( {{z}_{l}},{{z}_{r}},{{v}_{t}} \right), \\ 
 & z_{l}^{*}={{\gamma }_{l}}{{z}_{r\to l}}+{{z}_{l}}, \\ 
 & z_{r}^{*}={{\gamma }_{r}}{{z}_{l\to r}}+{{z}_{r}}, \\ 
    \end{aligned}
\end{equation}

\noindent where, ${\gamma }_{l}$ and ${\gamma }_{r}$ represent the weights for merging the left and right views, which are learnable variables. TASCA denotes the time-aware stereo ccross attention module, and its specific computation can be expressed in \cref{equ:3}.

\begin{equation}
\label{equ:3}
\begin{aligned}   
  & \text{TASC}{{\text{A}}_{r\to l}}=\text{GC(GC}(\text{TA}(W_{l}^{1}{{{\bar{z}}}_{l}},W_{r}^{1}{{{\bar{z}}}_{r}},W_{r}^{2}{{z}_{r}}))+{{W}^{v}}{{v}_{t}}) \\ 
 & \ \ \ \ \ \ \ \ \ \ \ \ \ \  +\text{GC}(\text{TA}(W_{l}^{1}{{{\bar{z}}}_{l}},W_{r}^{1}{{{\bar{z}}}_{r}},W_{l}^{2}{{z}_{r}})) \\ 
 & \text{TASC}{{\text{A}}_{l\to r}}=\text{GC(GC}(\text{TA}(W_{r}^{1}{{{\bar{z}}}_{r}},W_{l}^{1}{{{\bar{z}}}_{l}},W_{l}^{2}{{z}_{l}}))+{{W}^{v}}{{v}_{t}}) \\ 
 & \ \ \ \ \ \ \ \ \ \ \ \ \ \  +\text{GC}(\text{TA}(W_{r}^{1}{{{\bar{z}}}_{r}},W_{l}^{1}{{{\bar{z}}}_{l}},W_{l}^{2}{{z}_{l}})) \\ 
\end{aligned}
\end{equation}

\noindent where  $\overline{z}_{l} = LN(z_{l})$, $\overline{z}_{r} = LN(z_{r})$. $W_1^{l}$, $W_1^{r}$, $W_2^{l}$ and $W_2^{r}$ are projection matrices. The GC denotes a combination of group normalization \cite{wu2018group} and convolution layers, while $\operatorname{TA}$ refers to the temperature attention module, which can be represented by \cref{equ:4}

 \begin{equation}
 \label{equ:4}
    \operatorname{TA}(Q, K, V)=\operatorname{softmax}\left( \tau Q K^T / \sqrt{C} \right) V,
\end{equation}

 \noindent where $Q$, $K$, $V$ represent the input feature vector, and $C$ represents the dimension of the feature vector. $\tau$ is a hyperparameter representing the temperature coefficient.
 
\subsection{Training Loss}  
The training loss for DiffSteISR can be represented by \cref{equ:5}. Simply put, it predicts the noise $\epsilon$.

\begin{equation}
\label{equ:5}
\begin{aligned}
\mathcal{L} = \mathbb{E}_{z^{l}, z^{r}, t, \epsilon} \left[
\left\| \epsilon - \epsilon_\theta \left( z_t^{l}, z_t^{r}, z_{lr}^{l}, z_{lr}^{r}, t, p_{h}, p_{s} \right) 
\right\|_2^2
\right],
\end{aligned}
\end{equation}

\noindent where $z^{l}$ and $z^{r}$ denote the latent vectors of the HR images after VAE encoding, respectively. While $z_{lr}^{l}$ and $z_{lr}^{r}$ denote the latent vectors of the LR images after VAE encoding, respectively. The variable $t$ represents the randomly sampled diffusion steps. The variable $\epsilon$ is noise sampled from a standard normal distribution. The symbols $p_{h}$ and $p_{s}$ represent the hard tag prompt embedding and soft tag prompt embedding, respectively. The $z_t^{l}= \sqrt {{{\overline \alpha  }_t}} z^{l}+\sqrt {1 - {{\overline \alpha  }_t}}\epsilon $ and $z_t^{r}=\sqrt {{{\overline \alpha  }_t}} z^{r}+\sqrt {1 - {{\overline \alpha  }_t}}\epsilon$ denotes the left and right latent vectors after adding noise, respectively, and $\overline \alpha_t$ is a constant. The $\epsilon_\theta$ represents the DiffSteISR network model.

\section{Experiments And Analysis}

\subsection{Experimental Setup}

\textbf{Training Dataset:} Similar to previous works~\cite{zhou2024toward}, we used 800 pairs of high-resolution stereo images from the Flickr1024K dataset~\cite{wang2019flickr1024} and 60 pairs from the Middlebury dataset~\cite{scharstein2014high} as the training dataset. These images were cropped to fixed sizes of 512$\times$512 patches, which were then degraded to low-quality patches of the same size using the degradation method employed in RealSCGLAGAN~\cite{zhou2024toward}.

\textbf{Testing Dataset:} To compare with previous works, the synthetic dataset Flickr1024RS~\cite{zhou2024toward} and the real dataset StereoWeb20~\cite{zhou2024toward} were used to validate the performance of the proposed method.

\textbf{Implementation Details:} The SD-2.0 base model \footnote{https://huggingface.co/stabilityai/stable-diffusion-2} was utilized as the pre-trained diffusion prior model, and the Adam optimizer was used to optimize the SOA ControlNet and the inserted adapter layers during the diffusion process, with a batch size of 32 and a learning rate of 5e-5. The entire fine-tuning process was conducted on two NVIDIA A40 GPUs for a total of 100 epochs. For inference, we sampled 50 steps using the DDIM~\cite{song2020denoising} approach to ensure comparability with previous works.

\textbf{Evaluation Criteria:} To comprehensively evaluate the performance of different methods, a series of reference-based and no-reference metrics were employed to objectively assess the super-resolution quality on the synthetic dataset Flickr1024RS and the real dataset StereoWeb20. Specifically, for the Flickr1024RS dataset with GT, peak signal-to-noise ratio (PSNR) and structural similarity index measure (SSIM) were used to evaluate the fidelity, while LPIPS~\cite{zhang2018unreasonable} and DISTS~\cite{ding2020image} were utilized to assess perceptual quality. FID~\cite{heusel2017gans} was used to evaluate the distribution difference between the super-resolved images and the GT images. Additionally, mean absolute disparity error (MADE)~\cite{zhou2024toward} was introduced to assess the disparity consistency between the super-resolved images and GT images. For the no-GT StereoWeb20 dataset, we used NIQE~\cite{zhang2015feature}, MANIQA~\cite{yang2022maniqa}, MUSIQ~\cite{yang2022maniqa}, and CLIPIQA~\cite{yang2022maniqa} to evaluate the quality of the generated images.

\renewcommand{\arraystretch}{1.5} 

\begin{table*}[!htb]
\centering
\caption{Quantitative results achieved by different methods on the synthetic dataset Flickr1024RS \cite{zhou2024toward} and the real-world dataset StereoWeb20\cite{zhou2024toward} datasets. The best and second best results of each metric are highlighted in \textcolor{red}{red} and \textcolor{blue}{blue}.}
\label{tab:1}

\resizebox{\textwidth}{!}
{
\begin{tabular}{c|c|cccc|cccc}
\hline
\multicolumn{1}{c|}{Datasets} & Metrics & NAFSSR\cite{chu2022nafssr} & RealESRGAN~\cite{wang2021real} & HAT~\cite{chen2023activating} & RealSCGLAGAN~\cite{zhou2024toward} & StableSR~\cite{wang2024exploiting} & PASD~\cite{yang2023pixel} & SeeSR~\cite{wu2024seesr} & DiffSteISR(Ours) \\ \hline
 & PSNR$\uparrow$ & {20.80} & 20.50 & \textcolor{blue}{\textbf{20.90}} & \textcolor{red}{\textbf{21.06}} & 20.46 & 20.41 & 20.46 & 20.13 \\
 & SSIM$\uparrow$ & 0.5696 & {0.5714} & \textcolor{blue}{\textbf{0.5780}} & \textcolor{red}{ \textbf{0.6235}} & 0.5427 & 0.5206 & 0.5263 & 0.5089 \\
 & MADE$\downarrow$ & 4.7525 & {3.6461} & \textcolor{blue}{ \textbf{3.3389}} & \textcolor{red}{ \textbf{2.1988}} & 4.8660 & 8.5305 & 6.9405 & 3.9298 \\
 & LPIPS$\downarrow$ & 0.4588 & \textcolor{blue}{ \textbf{0.3001}} & 0.3198 & \textcolor{red}{ \textbf{0.2098}} & 0.3060 & 0.3251 & {0.3010} & 0.3248 \\
 & DISTS$\downarrow$ & 0.2264 & 0.1631 & 0.1790 & \textcolor{red}{ \textbf{0.1148}} & {0.1562} & \textcolor{blue}{ \textbf{0.1550}} & 0.1693 & 0.1672 \\
 & FID$\downarrow$ & {58.56} & 62.03 & 62.87 & \textcolor{red}{ \textbf{40.20}} & 61.74 & 62.31 & \textcolor{blue}{ \textbf{55.31}} & 64.02 \\ \cline{2-10} 
 & NIQE$\downarrow$ & 5.9410 & \textcolor{red}{ \textbf{3.2429}} & 3.9595 & {3.4737} & 3.6912 & \textcolor{blue}{\textbf{3.4685}} & 3.5344 & 3.7246 \\
 & MANIQA$\uparrow$ & 0.5258 & 0.6092 & 0.6071 & \textcolor{blue}{ \textbf{0.6555}} & 0.6349 & 0.6423 & {0.6507} & \textcolor{red}{ \textbf{0.6584}} \\
 & MUSIQ$\uparrow$ & 53.43 & 67.33 & 63.08 & \textcolor{red}{\textbf{71.52}} & 67.59 & 70.58 & { 70.60} & \textcolor{blue}{ \textbf{71.43}} \\
\multirow{-10}{*}{\begin{tabular}[c]{@{}c@{}}Flickr1024RS\\ (synthetic dataset)\end{tabular}} & CLIPIQA$\uparrow$ & 0.3965 & 0.5236 & 0.4303 & 0.6544 & 0.5996 & \textcolor{red}{ \textbf{0.6750}} & 0.6543 & \textcolor{blue}{ \textbf{0.6702}} \\ \hline
 & NIQE$\downarrow$ & 5.7363 & \textcolor{blue}{ \textbf{4.1916}} & 4.6925 & {4.4174} & 4.5196 & \textcolor{red}{ \textbf{3.8492}} & {4.2945} & 4.7570 \\
 & MANIQA$\uparrow$ & 0.4648 & 0.5687 & 0.5657 & 0.5761 & 0.5895 & {0.6223} & \textcolor{blue}{ \textbf{0.6268}} & \textcolor{red}{\textbf{0.6314}} \\
 & MUSIQ$\uparrow$ & 46.78 & 60.76 & 57.96 & 62.18 & 59.91 & {64.85} & \textcolor{blue}{\textbf{65.06}} & \textcolor{red}{\textbf{66.01}} \\
\multirow{-4}{*}{\begin{tabular}[c]{@{}c@{}}StereoWeb20\\ (real-world dataset)\end{tabular}} & CLIPIQA$\uparrow$ & 0.5144 & 0.5414 & 0.4622 & 0.6331 & 0.5966 & \textcolor{blue}{\textbf{0.6945}} & {0.6846} & \textcolor{red}{\textbf{0.6985}} \\ \hline
\end{tabular}}

\vspace{-3mm}
\end{table*}

\subsection{Comparison with State-of-the-Art Methods}

We compared DiffSteISR with current state-of-the-art methods in stereo super-resolution. Due to the limited availability of relevant real-world stereo super-resolution algorithms, we selected representative methods such as NAFSSR~\cite{chu2022nafssr} and RealSCGLAGAN~\cite{zhou2024toward}. Additionally, single-image super-resolution (SISR) methods, including GAN-based algorithms like RealESRGAN~\cite{wang2021real} and HAT~\cite{chen2023activating}, as well as DM-based methods such as StableSR~\cite{wang2024exploiting}, PASD~\cite{yang2023pixel}, and SeeSR~\cite{wu2024seesr}, were included for comparison.

\textbf{Quantitative Evaluation:} \cref{tab:1} presents the quantitative comparison results on the synthetic dataset Flickr1024RS and the real dataset StereoWeb20. From the table, the following conclusions were drawn : (1) Compared to traditional GAN-based super-resolution methods, DM-based methods achieve better results on no-reference metrics (NIQE, MANIQA, MUSIQ, and CLIPIQA). In particular, DiffSteISR achieved competitive results among DM methods, especially on the real dataset StereoWeb20, confirming the effectiveness and advantages of DM-based methods; (2) Compared to DM-based methods, traditional GAN methods have a significant advantage in pixel-level metrics such as PSNR and SSIM. This is primarily because GAN-base methods introduce pixel-level loss during training, optimizing the pixel differences between input data and GT data. In contrast, DM-based methods model the real data distribution, leading to outputs that tend to generate realistic and natural texture details, sacrificing some pixel-level consistency; (3) Although perceptual metrics are calculated at the feature level, GAN-base methods tend to achieve higher precision than DM-based methods on perceptual quality metrics like LPIPS and DISTS. This is mainly due to diffusion models generating richer details, leading to outputs that are not fully consistent with GT, thus resulting in certain feature discrepancies; (4) In terms of disparity consistency measured by MADE, GAN-based methods generally outperform DM-based methods, primarily because the details generated by diffusion-based methods are not fully consistent with GT, leading to noticeable disparities when calculating depth. Notably, the proposed method achieved a MADE of 3.9298, indicating that it maintains better disparity consistency compared to other DM-based methods. Overall, our method demonstrates competitive results on no-reference metrics while maintaining high disparity consistency with GT.
\begin{figure*}
    \centering
    \includegraphics [width=0.95\textwidth]{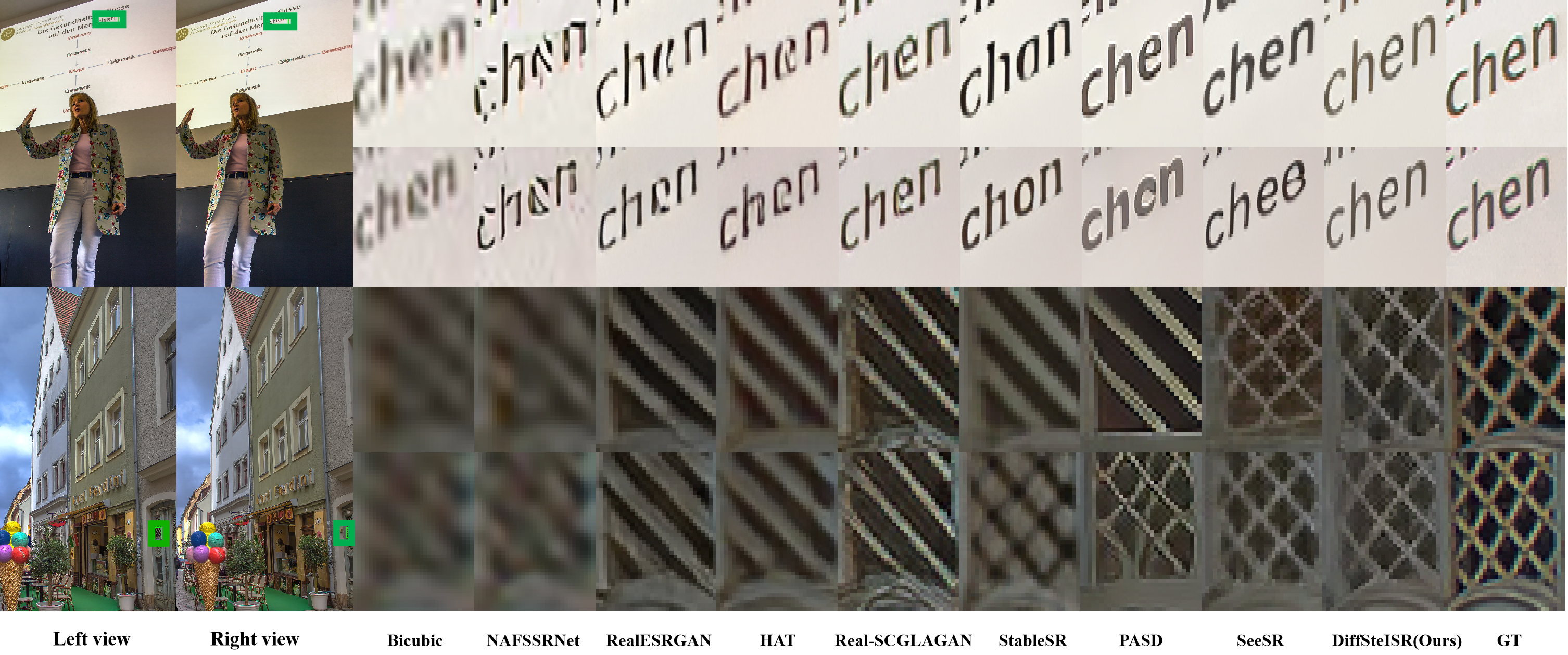}
    \caption{Visual results ($\times$4) achieved by different methods on the Flickr1024RS~\cite{zhou2024toward} dataset.}
    \label{fig:6}
\end{figure*}
\begin{figure*}
    \centering
    \includegraphics [width=0.95\textwidth]{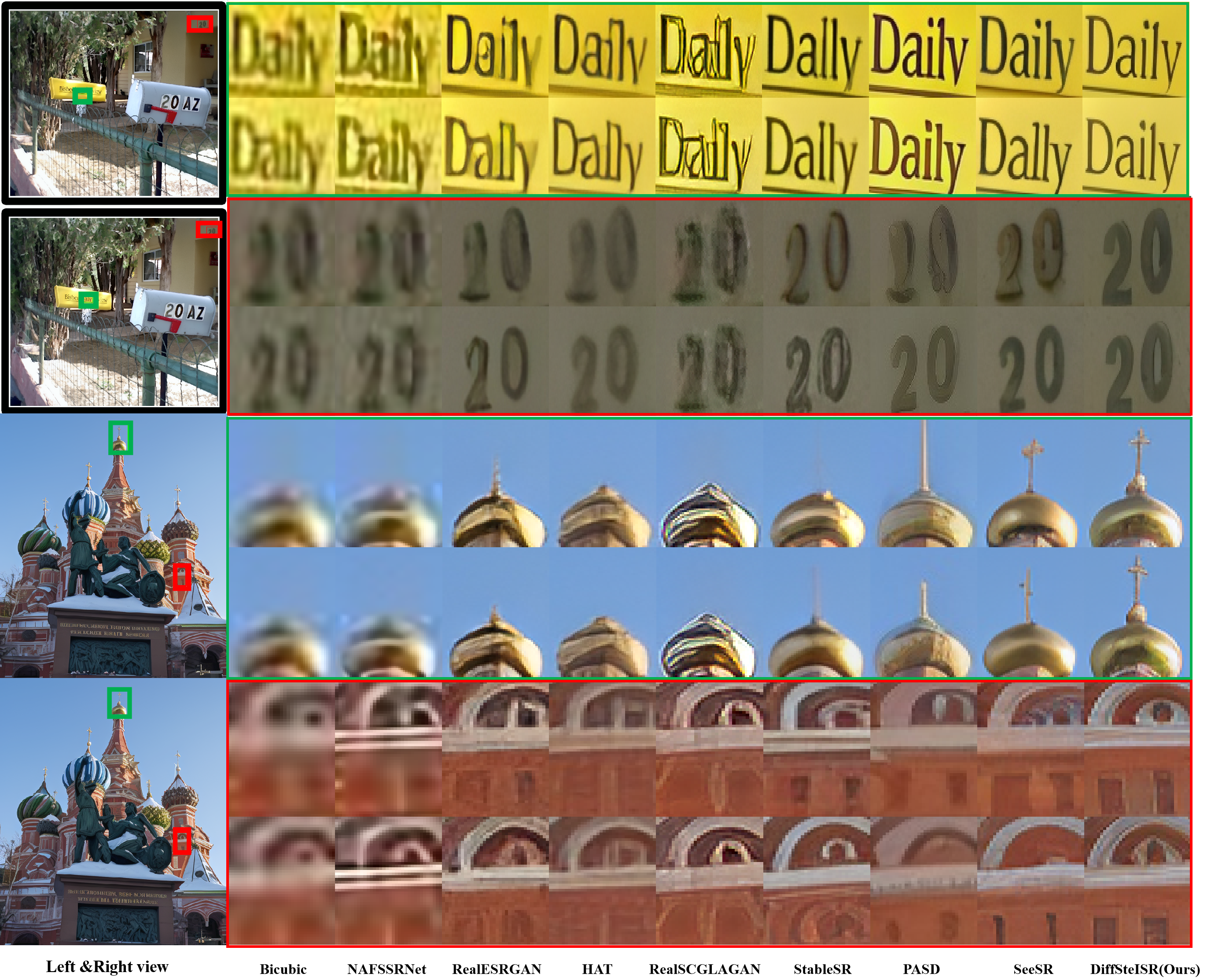}
    \caption{Visual results ($\times$4) achieved by different methods on the StereoWeb20~\cite{zhou2024toward} dataset.}
    \label{fig:7}
\end{figure*}

\textbf{Qualitative Evaluation:} \cref{fig:6} shows the qualitative evaluation results on the synthetic dataset Flickr1024RS. In the reconstruction of the word ``chen", DM-based methods are able to recover clearer boundaries and accurate characters. In the reconstruction of the fence, only DM-based methods could recreate the grid-like realistic texture, while GAN-based methods typically produced spurious diagonal patterns. This highlights the superior capability of diffusion models in reconstruction severely degraded images compared to GAN-based methods. However, it is worth noting that SISR methods based on diffusion model (DM), such as StableSR~\cite{wang2024exploiting}, PASD~\cite{yang2023pixel}, and SeeSR~\cite{wu2024seesr}, lack the consistency constraints of left and right views, leading to discrepancies in texture details between the generated left and right images. For example, StableSR~\cite{wang2024exploiting} and PASD~\cite{yang2023pixel} exhibit significant texture differences in their left and right images, while SeeSR~\cite{wu2024seesr} generates similar grid textures, but with noticeable size differences. In contrast, our proposed DiffSteISR not only recovers realistic and natural textures but also maintains high consistency in texture between the left and right views, effectively demonstrating the validity of our method.

\cref{fig:7} further showcases the qualitative evaluation on the real dataset StereoWeb20. It is evident from the figure that DM-based methods tend to reconstruct realistic and natural textures, whereas GAN-based methods often suffer from severe artifacts and false textures, particularly noticeable in letters and numbers. These examples strongly confirm the effectiveness of -DM-based super-resolution methods. Notably, among DM-based methods, DiffSteISR achieves better visual results, with more reasonable texture generation and high consistency between the left and right views, effectively demonstrating its advantages in the field of real-world stereo super-resolution.
\begin{figure*}
    \centering
    \includegraphics [width=0.95\textwidth]{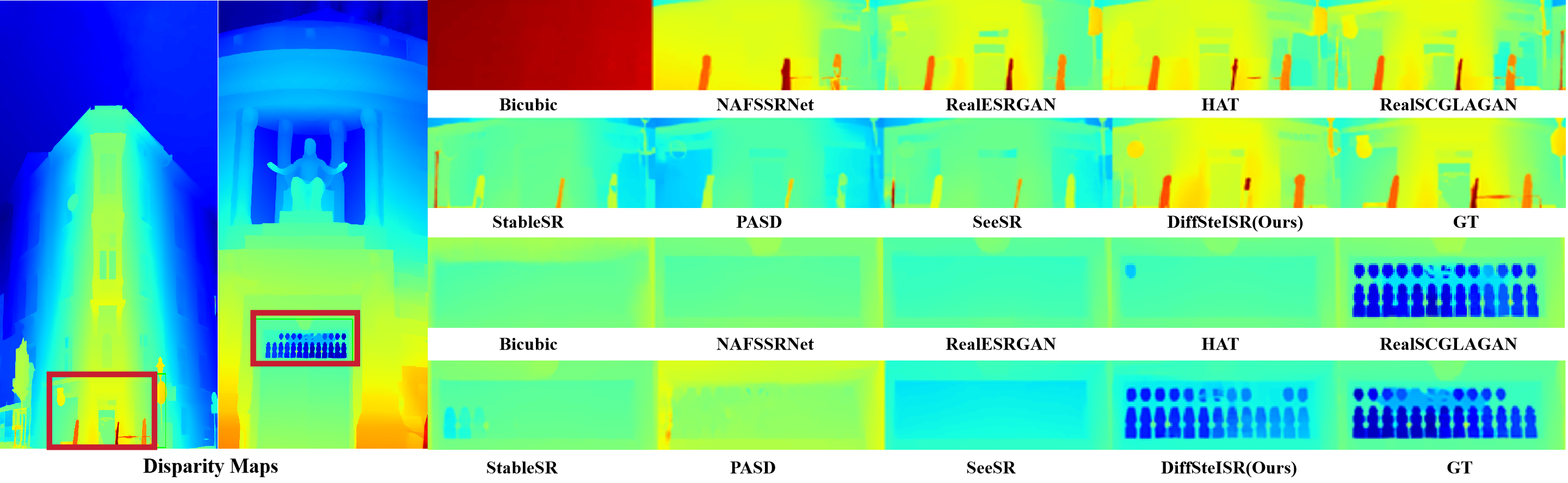}
    \caption{The visual results of disparity estimated images achieved by different methods on the Flickr1024RS~\cite{zhou2024toward} dataset.}
    \label{fig:8}
\end{figure*}
\cref{fig:8} qualitatively compares the impact of different methods on disparity after enhancement. It can be observed that GAN-based super-resolution methods generally maintain high disparity consistency with GT, while DM-based super-resolution methods perform poorly due to the inherent randomness, which leads to lower consistency in texture between the left and right views, as illustrated in \cref{fig:6} and \cref{fig:7}. Notably, our proposed DiffSteISR effectively extends the application of DMs in the field of real-world stereo super-resolution by introducing a series of dual-view fusion techniques, including SSE, SOA ControlNet and TASCATA.

\subsection{User Study}
\begin{figure}
    \centering
    \includegraphics [width=0.5\textwidth]{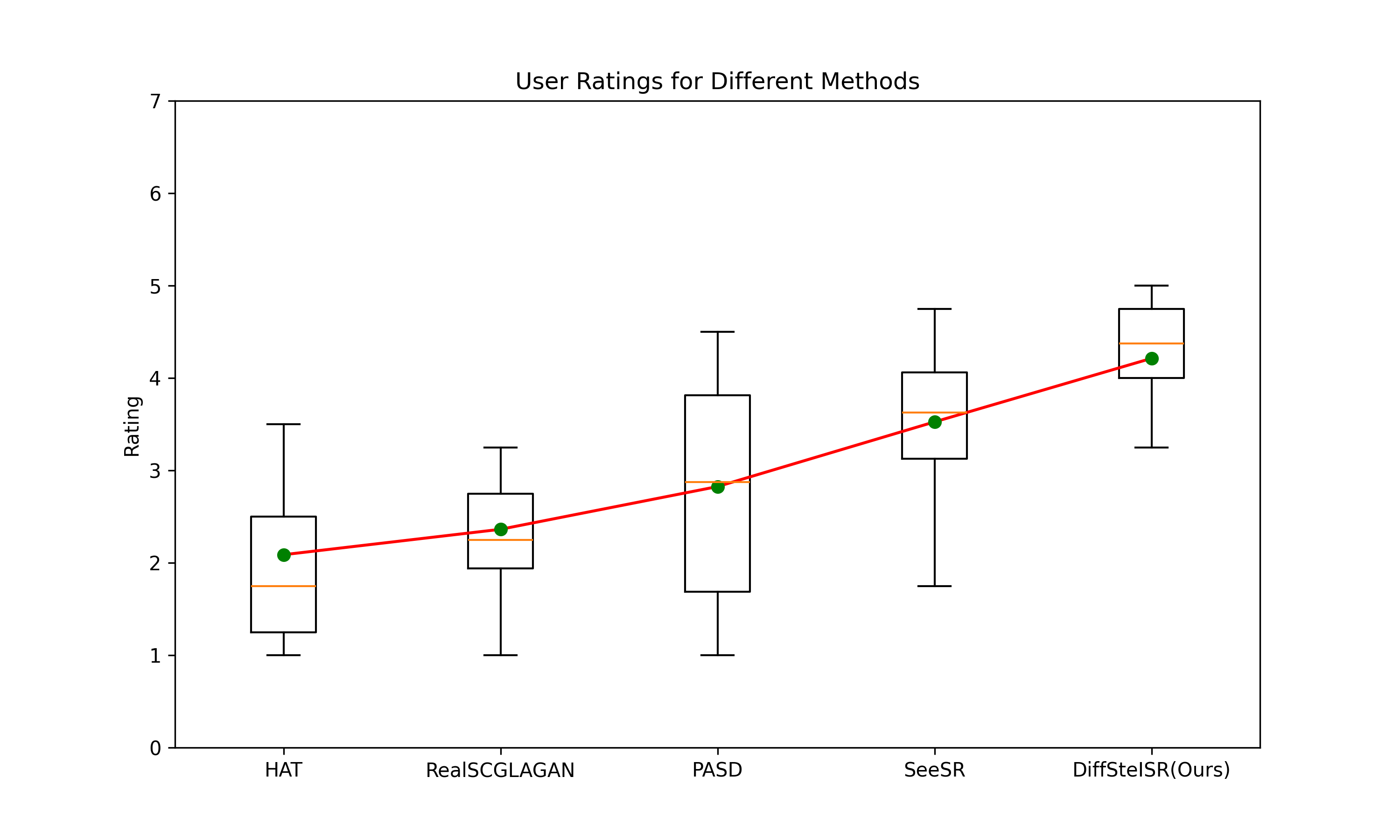}
    \caption{Results of user study on real-world data.}
    \label{fig:9}
\end{figure}
To further validate the effectiveness of the proposed method, we collected 40 real low-resolution stereo images (20 pairs) and conducted a user study with 20 participants. The study compared methods that performed well in previous qualitative and quantitative evaluations, including HAT~\cite{chen2023activating}, RealSCGLAGAN~\cite{zhou2024toward}, PASD~\cite{yang2023pixel}, SeeSR~\cite{wu2024seesr}, and DiffSteISR. Participants were asked to evaluate the five methods based on perceived image quality, semantic correctness of the LR images, and the consistency of texture between the left and right images. The scoring was on a scale of 1 to 5, where 5 represented the best quality and 1 the worst.

After averaging the scores from the 20 participants, the results, shown in \cref{fig:9}, indicate that methods based on diffusion models generally yield significantly better subjective visual quality compared to GAN-based methods. However, PASD exhibited a wide interquartile range (IQR), indicating variability in participant evaluations. In contrast, DiffSteISR not only achieved the highest median score but also had a narrower IQR, suggesting that the majority of users rated it highly, thereby confirming the effectiveness of the proposed algorithm.

\subsection{Ablation Study}
\begin{figure*}
    \centering
    \includegraphics [width=0.95\textwidth]{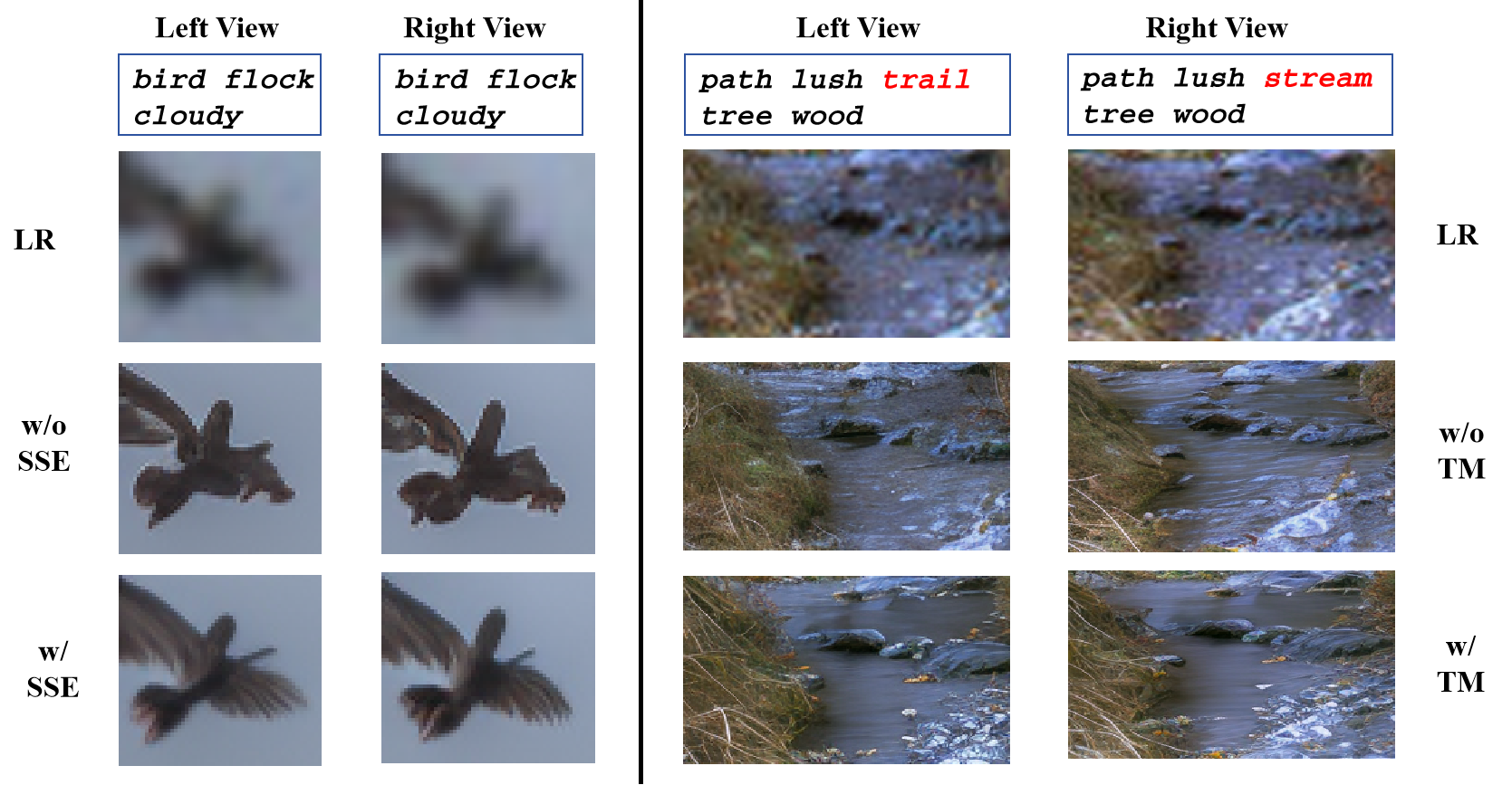}
    \caption{The results on the ablation of stereo semantic extractor and its tag merging module.}
    \label{fig:10}
\end{figure*}
\begin{figure}
    \centering
    \includegraphics [width=0.45\textwidth]{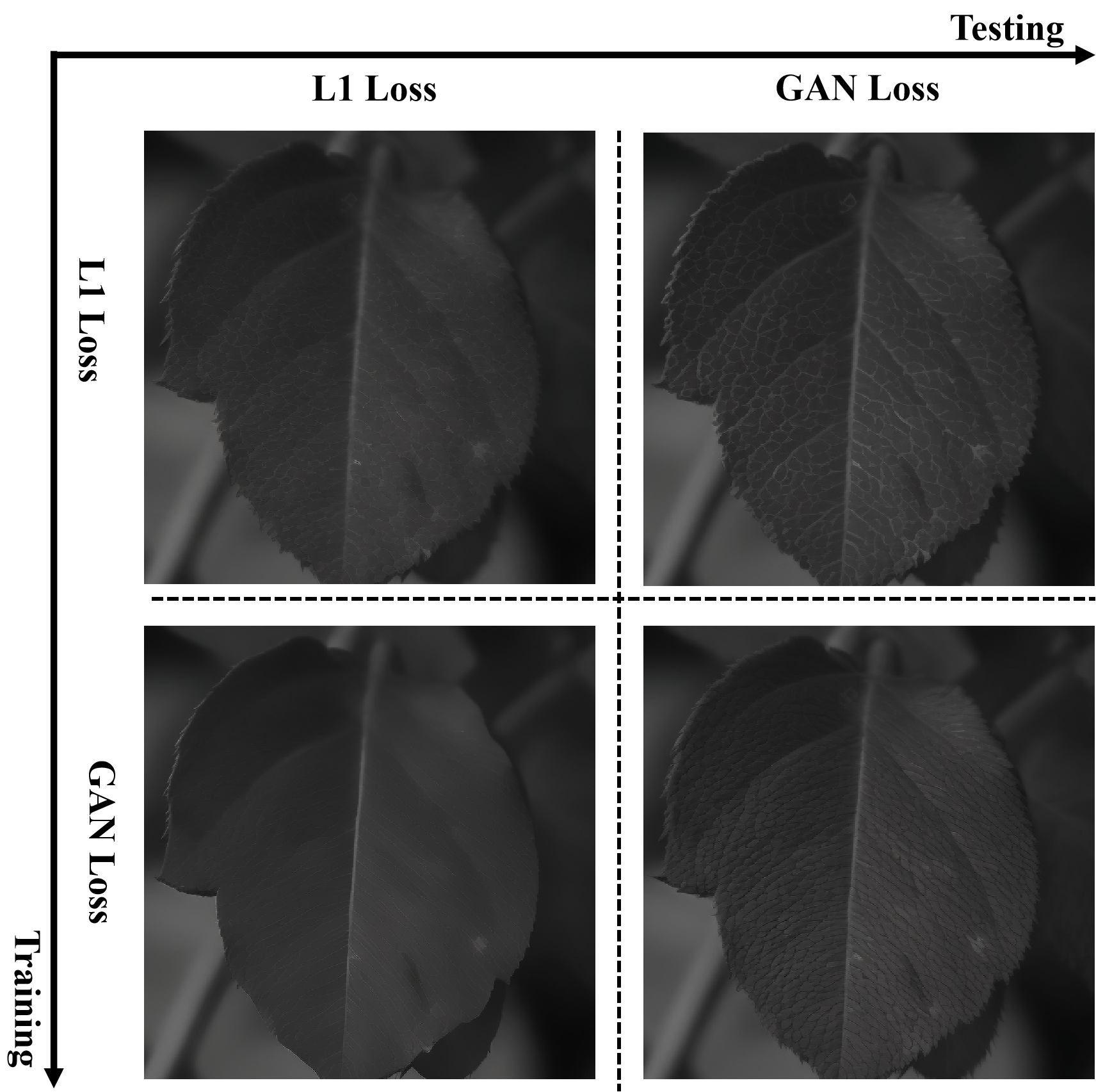}
    \caption{The ablation results on the impact of using different pre-training losses on generating image textures for stereo omni attention network in SOA ControlNet.}
    \label{fig:11}
\end{figure}
\begin{figure*}
    \centering
    \includegraphics [width=0.95\textwidth]{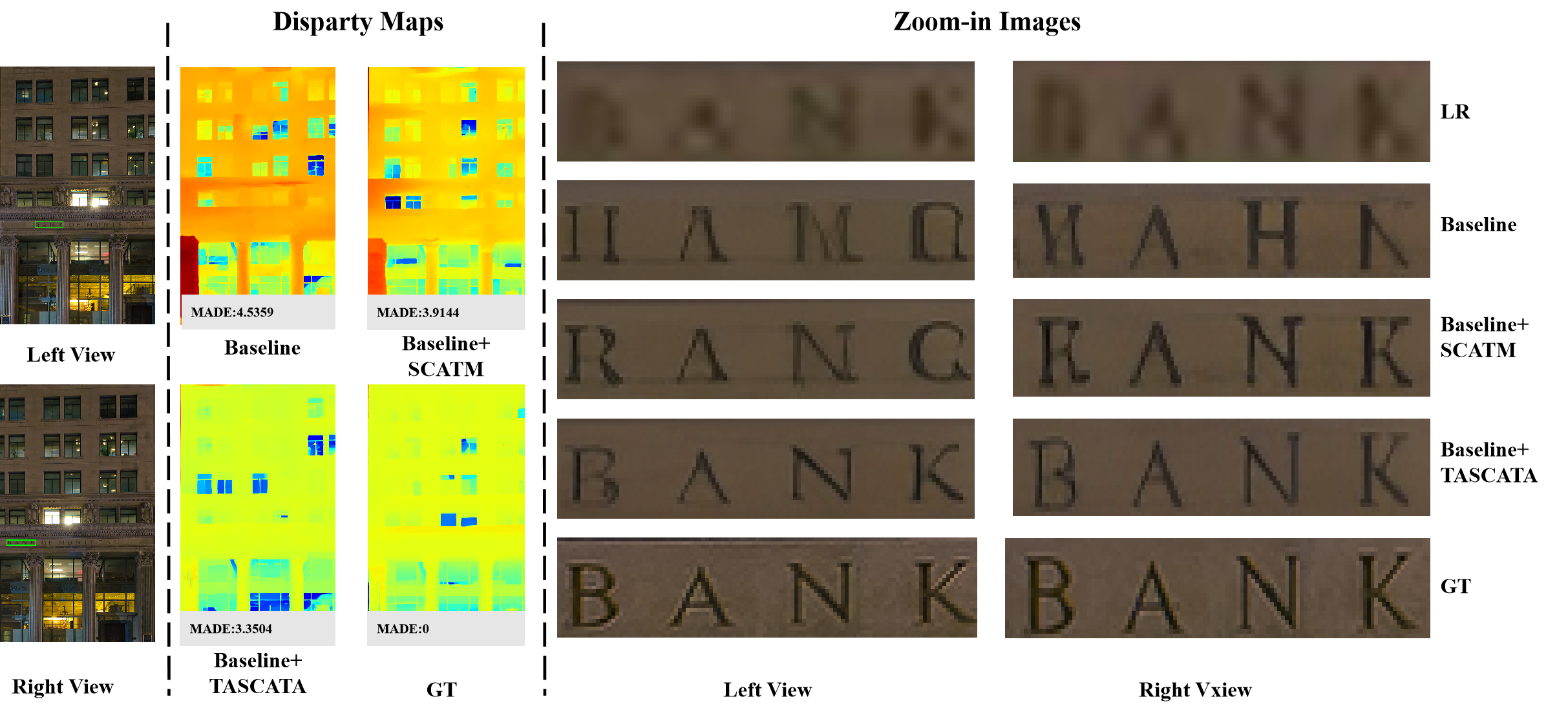}
    \caption{The ablation results with different stereo fusion modules.}
    \label{fig:12}
\end{figure*}
\textbf{Effectiveness of the Stereo Semantic Extractor (SSE):} \cref{fig:10} (left) shows a visual comparison of generated images before and after incorporating the SSE module. The clarity of the bird's feathers was notably improved after the addition of the SSE module, confirming its effectiveness. \cref{fig:10} (right) illustrates the visual comparison of generated images before and after the implementation of the tag merging (TM) strategy. Without the TM strategy, the left and right images exhibited semantic differences due to varying prompts. After introducing the TM strategy, the consistency between the generated left and right images improved significantly, demonstrating both the effectiveness and necessity of the TM strategy.

\textbf{Effectiveness of the SOA ControlNet:} To validate the effectiveness of the SOA ControlNet, we performed experiments on the Flickr1024RS dataset under the following conditions: (1) no ControlNet (baseline); (2) original ControlNet; and (3) the proposed SOA ControlNet. \cref{tab:2} presents the quantitative evaluation results. It is evident that adding the ControlNet significantly improved the reference metrics (PSNR, MADE, LPIPS, FID) and the no-reference evaluation metric CLIPIQA, indicating that the addition of ControlNet enhances the effectiveness of the generated images in terms of pixel-level, perceptual-level, and distribution-level consistency. Furthermore, the use of the proposed SOA ControlNet led to significant improvements in CLIPIQA and further reductions in MADE, proving its effectiveness in enhancing visual quality and reducing disparity error.

 \begin{table}
\centering
\caption{Ablation results achieved on Flickr1024RS trained with different ControlNet modules.}
\label{tab:2}
\resizebox{0.5\textwidth}{!}
{
\begin{tabular}{llllll}\hline
                          & \multicolumn{5}{c}{Metrics}                                     \\ \cline{2-6} \multirow{-2}{*}{Methods}  & PSNR$\uparrow$  & MADE$\downarrow$    & LPIPS$\downarrow$   & FID$\downarrow$    & CLIPIQA$\uparrow$                        \\ \hline
(1) baseline                  & 19.49 & 7.6355 & 0.3520 & 65.07 & 0.5646                        \\
(2) baseline+ControlNet       & 19.98 & 6.4626 & 0.3309 & 64.34 & 0.5809                        \\                    
(3) baseline+SOA ControlNet   & 20.23 & 3.9476 & 0.3266 & 64.01 & {\color[HTML]{000000} 0.6641} \\ \hline\end{tabular}}
\vspace{-2mm}
\end{table}

Additionally, as described in~\cref{sec:soacn}, the SOAN in SOA ControlNet is a pre-trained model whose parameters remain unchanged during both training and inference of DiffSteISR. The L1 loss is most straightforward approach to train SOAN, as suggested by PASD~\cite{yang2023pixel}. However, we have observed an interesting phenomenon: using an SOAN pre-trained with L1 loss during the training of DiffSteISR, and an SOAN pre-trained with GAN Loss during the inference phase, can effectively enhance the visual quality of the generated images. Therefore, this paper investigates the impact of employing different loss functions to train the SOAN during the training and inference stages of DiffSteISR on the visual quality of the generated images. The final qualitative comparison results are shown in \cref{fig:11}. We hypothesize that using an SOAN trained with L1 Loss during DiffSteISR training leads to smoother inputs for the diffusion model, which in turn encourages the diffusion model to generate richer textures. Subsequently, if an SOAN trained with GAN loss is used during the inference phase of DiffSteISR, the input textures provided to the diffusion model are preserved more effectively than those from L1 loss training. Consequently, under the prior of the diffusion model's tendency to generate more textures, the diffusion model is able to produce images with richer textures.

\textbf{Effectiveness of the TASCATA:} To explore the effectiveness of the TASCATA, we conducted studies on the Flickr1024RS dataset under the following conditions: (1) no stereo information fusion adapter; (2) SCATM as the stereo information fusion adapter; and (3) the TASCATA as the stereo information fusion adapter. The quantitative evaluation results are shown in \cref{tab:3}. Comparing experiments (1) and (2)/(3) reveals a significant reduction in MADE after incorporating the stereo information fusion adapter, demonstrating its necessity in stereo image super-resolution based on DM. However, we also observed a decrease in the non-reference metric CLIPIQA. We believe that the inclusion of the stereo information fusion module causes the model to prioritize the relationship between the left and right images, thereby sacrificing the generation of richer textures. The comparison between experiments (2) and (3) shows that the proposed TASCATA further reduces the MADE value by 1.4059. Moreover, it partially compensates for the loss in image quality introduced by the inclusion of the stereo information fusion module, providing strong evidence for its effectiveness.

 \begin{table}
\centering
\caption{Ablation results achieved on Flickr1024RS trained with different stereo fusion modules.}
\label{tab:3}
\resizebox{0.5\textwidth}{!}
{
\begin{tabular}{llllll}\hline\multirow{2}{*}{Methods} & \multicolumn{5}{c}{Metrics}               \\ \cline{2-6} 
                        & PSNR$\uparrow$  & MADE$\downarrow$    & LPIPS$\downarrow$   & FID$\downarrow$    & CLIPIQA$\uparrow$ \\ \hline
(1) baseline                & 20.04 & 8.7969 & 0.3183 & 63.39 & 0.6719  \\
(2) baseline+SCATM          & 20.12 & 5.4168 & 0.3167 & 63.21 & 0.6562  \\
(3) baseline+TASCATA        & 20.10 & 4.0109 & 0.3130 & 63.48 & 0.6699  \\ \hline\end{tabular}}
\vspace{-2mm}
\end{table}
\cref{fig:12} further shows the qualitative evaluation results. Firstly, the disparity maps indicate that the incorporation of TASCATA significantly improves the disparity consistency between the enhanced images and GT images. Secondly, by examining the zoom-in images, it is apparent that models without any stereo information fusion modules struggle to recover the English characters in the corresponding left and right images, while those with the streo information fusion module exhibit notable improvements in texture and structural consistency. The proposed TASCATA is particularly effective in enhancing the consistency of texture between the left and right images.

\subsection{Discussion}
  Compared to traditional GAN-based  methods, DiffSteISR introduces the DM to effectively reconstruct more natural and realistic textures. Another significant benefit is the reduced need for meticulous adjustments of the discriminator and generator structures, as well as the training loss weights. However, disparity in stereo images is a crucial aspect in practical applications, as lower disparity errors lead to more accurate depth modeling.
  
In our research, we observed a considerable gap between GAN-based and DM-based methods regarding the reduction of disparity errors. Despite our efforts to bridge this gap, DiffSteISR still exhibits certain limitations in achieving parity with GAN-base methods. Therefore, exploring ways to further minimize disparity errors in DM-based stereo image super-resolution remains an important avenue for future research.

\section{Conclusion}
This paper presents DiffSteISR, a pioneering DM-based approach for  real-world stereo images super-resolution. By mastering the diffusion priors, the TASCATA, the SOA ControlNet, and the SSE, DiffSteISR effectively reconstruct the loss details of low-resolution stereo images while ensuring high consistency and accuracy in texture and semantics between the left and right views. Extensive experimental results demonstrate that DiffSteISR produces more realistic and natural textures compared to GAN-based methods while exhibits improved disparity alignment with GT images compared to DM-based single-image super-resolution models, which reveals the strong competitiveness. We believe that our method provides a solid foundation for future research in the field.
\section*{Acknowledgements}
This work was supported by National Natural Science Foundation of China under Grant 62171133, in part by the Fujian Health Commission under Grant 2022ZD01003.

\bibliographystyle{ieeetr}
\bibliography{main}

\end{document}